\documentclass[11pt]{article}

\usepackage[utf8]{inputenc}
\usepackage[T1]{fontenc}
\usepackage{lmodern}
\usepackage{geometry}
\usepackage{hyperref}
\usepackage{authblk}
\usepackage{orcidlink}

\title{Evaluating Multilingual Sentence Embeddings for Translation Error Detection:
An English--Greek Contrastive Study}

\author[1]{Eleftherios Kalogeros \orcidlink{0000-0001-8080-6343}}
\author[1]{Athanasios Ntalakas \orcidlink{0009-0001-7098-6296}}
\author[1]{Manolis Gergatsoulis \orcidlink{0000-0001-9744-5839}}
\author[2]{Paschalis Nikolaou \orcidlink{0000-0001-6839-4999}}
\author[2]{Sotiria-Lito Alexaki \orcidlink{0009-0007-6331-4102}}

\affil[1]{%
Laboratory on Digital Libraries and Electronic Publishing\\
Department of Information Science, Ionian University\\
Ioannou Theotoki 72, 49100 Corfu, Greece\\
\texttt{kalogero@ionio.gr, at.ntalakas@gmail.com, manolis@ionio.gr}
}

\affil[2]{%
Centre for Literary Translation and Book Policy\\
Department of Foreign Languages, Translation and Interpreting,
Ionian University\\
Tsirigoti Sq., 49100 Corfu, Greece\\
\texttt{nikolaou@ionio.gr, lito\_alexaki@ionio.gr}
}

\date{}

\begin{document}

\maketitle

\begin{abstract}
\emph{Multilingual sentence embeddings} are increasingly used to estimate semantic
similarity across languages, yet their sensitivity to fine-grained translation
errors remains insufficiently understood. This study investigates whether
general-purpose multilingual embedding models can distinguish correct
English--Greek translations from minimally modified erroneous alternatives.
A contrastive dataset was developed from FLORES+ sentence-aligned reference
translations and double-reviewed by two translation experts. It contains
1,850 examples distributed across ten core and five exploratory error
categories, covering factual, lexical-semantic, grammatical, relational,
referential, and discourse-level phenomena.

Five multilingual sentence-embedding models---BGE-M3, Multilingual E5,
Multilingual MPNet, LaBSE, and Jina Embeddings v3---were evaluated using
\emph{cosine similarity} between each English source sentence and its correct and
erroneous Greek translations. A reference-free COMETKiwi quality-estimation
model was additionally evaluated as a dedicated machine translation baseline.
Performance was assessed primarily through \emph{contrastive accuracy}, with
within-model score margins used to analyze \emph{category-specific sensitivity}.
BGE-M3 achieved the highest contrastive accuracy among the embedding models
at 89.30\%, whereas COMETKiwi achieved 94.49\%. Across the five embedding
models, explicit factual and lexical changes were detected more reliably than
tense-and-aspect and pronoun--coreference errors. COMETKiwi substantially
improved performance on several of these difficult categories, including tense
and aspect, pronoun and coreference, and semantic-role errors, but showed
markedly lower sensitivity to date-and-time errors and also underperformed the
embedding models on numbers. These findings indicate that general-purpose
multilingual embeddings and MT-specific quality estimation exhibit
complementary error-sensitivity profiles. Multilingual sentence embeddings
therefore provide useful semantic adequacy signals but are better suited as
components of broader translation-evaluation frameworks than as standalone
metrics.
\end{abstract}


\section{Introduction}
\label{sec:introduction}

Automatic evaluation plays a central role in the development and comparison of
machine translation systems. Traditional \emph{lexical metrics}, such as BLEU,
measure surface-level overlap between a candidate translation and one or more
reference translations~\cite{papineni-etal-2002-bleu}, but such metrics may
assign low scores to valid translations that differ lexically from the
reference. More recent evaluation approaches therefore incorporate contextual
and cross-lingual pretrained representations in order to assess translation
quality at a more semantic level~\cite{rei-etal-2020-comet}.

In parallel, general-purpose \emph{multilingual sentence-embedding models} have been
developed for applications such as cross-lingual retrieval, semantic search,
sentence alignment, and similarity estimation~\cite{chen-etal-2024-m3}.
These models aim to map semantically related sentences from different
languages into a shared embedding space, making them potentially useful for
translation evaluation.

Despite their strong performance on broad semantic-similarity and retrieval
tasks, it remains unclear whether multilingual sentence embeddings are
sufficiently sensitive to fine-grained translation errors. A correct
translation and a translation containing a single critical error may share
most of their lexical content and differ only in a number, a negation marker,
a pronoun, a temporal expression, or the semantic roles of two participants.
Such sentences may therefore occupy very similar positions in the embedding
space, even though their meanings are not equivalent. High cosine similarity
alone may consequently conceal translation errors that substantially change
the factual, grammatical, referential, or discourse-level interpretation of a
sentence.

This study investigates the extent to which multilingual sentence embeddings
can distinguish correct English--Greek translations from minimally modified
erroneous alternatives. English--Greek provides a useful setting for this
analysis because Greek exhibits rich morphology, grammatical gender, overt
agreement, flexible word order, and discourse-dependent reference.

A contrastive English--Greek dataset was constructed from sentence-aligned
FLORES+ reference translations. It contains 1,850 examples in which each
English source sentence is paired with a correct Greek translation and a
minimally modified erroneous alternative. The dataset covers 15 translation
error categories spanning factual, lexical-semantic, grammatical, relational,
referential, and discourse-level phenomena.

We evaluate five publicly available multilingual sentence-embedding models:
BGE-M3~\cite{chen-etal-2024-m3}, Multilingual E5~\cite{wang2024multilingual},
Multilingual MPNet~\cite{reimers-gurevych-2020-making},
LaBSE~\cite{feng-etal-2022-language}, and Jina Embeddings
v3~\cite{sturua2024jina}. To contextualize their performance against a
translation-specific evaluation approach, we additionally evaluate a
reference-free COMETKiwi quality-estimation model. Model performance is
assessed using contrastive accuracy and score margins between the correct and
erroneous translations, with category-level analysis used to identify shared
and method-specific strengths and weaknesses.

The study addresses the following research questions:

\begin{description}

\item[\textbf{RQ1:}] To what extent can multilingual sentence-embedding
models assign higher cosine similarity to correct English--Greek translations
than to translations containing controlled errors?

\item[\textbf{RQ2:}] How does error-detection sensitivity vary across
factual, lexical-semantic, grammatical, relational, referential, and
discourse-level translation-error categories?

\item[\textbf{RQ3:}] How do BGE-M3, Multilingual E5, Multilingual MPNet, LaBSE,
and Jina Embeddings v3 differ in contrastive accuracy and in their within-model
similarity-margin patterns between correct and erroneous translations?

\item[\textbf{RQ4:}] Which translation-error categories constitute shared or
model-specific blind spots for general-purpose multilingual sentence
embeddings?

\item[\textbf{RQ5:}] Can multilingual sentence embeddings be used as
standalone translation-evaluation metrics, or are they better suited as
semantic components within broader evaluation frameworks?

\end{description}

The main contribution of this work is a category-based empirical analysis of
multilingual sentence embeddings for English--Greek translation evaluation.
Rather than reporting only aggregate model performance, the study examines the
specific semantic distinctions preserved or lost in the embedding spaces of
five models with different training objectives. It additionally contributes a
human-reviewed contrastive challenge set covering a broad range of factual,
lexical, grammatical, referential, relational, and discourse-level errors. The
results provide evidence concerning the practical usefulness of sentence
embeddings as translation-evaluation signals while also identifying the
linguistic phenomena for which additional lexical, syntactic, referential,
inference-based, or discourse-aware modelling is required.

The remainder of this paper is organized as follows.
Section~\ref{sec:sentence-embedding} introduces the multilingual sentence-embedding models evaluated in
the study, and Section~\ref{sec:related-work} reviews the related work based on translation-error analysis,
contrastive evaluation. Section~\ref{sec:methodology} presents
the contrastive evaluation methodology, while Section~\ref{sec:experimental-setup} describes the model
configurations and COMETKiwi baseline. Section~\ref{sec:dataset-construction} details the construction of
the English--Greek contrastive dataset. Section~\ref{sec:results} reports the overall and
category-level experimental results, which are discussed in Section~\ref{sec:discussion}.
Finally, Section~\ref{sec:conclusion} summarizes the main findings and outlines directions for
future work.

\section{Preliminaries: Multilingual Sentence-Embedding Models}
\label{sec:sentence-embedding}

\emph{Sentence embeddings} represent sentences as dense numerical vectors in a semantic space, where sentences with similar meanings are expected to be located close to one another. Their similarity is commonly measured using \emph{cosine similarity}. \emph{Multilingual sentence-embedding models} extend this idea across languages by mapping semantically equivalent sentences into a shared vector space.

Five multilingual sentence-embedding models are examined in the present study: \emph{BGE-M3}, \emph{Multilingual E5}, \emph{Multilingual MPNet}, \emph{LaBSE}, and \emph{Jina
Embeddings v3} (Table~\ref{tab:model-setup}). The corresponding checkpoints are publicly available through the Hugging Face model repository\footnote{\url{https://huggingface.co/models}} and can be executed locally.

\emph{BGE-M3}~\cite{chen-etal-2024-m3} is a multilingual, multi-functional, and multi-granularity embedding model supporting dense, sparse, and multi-vector representations. It was developed primarily for multilingual and cross-lingual retrieval and supports more than 100 languages.

\emph{Multilingual E5}~\cite{wang2024multilingual} is a family of multilingual embedding models trained through large-scale contrastive learning. It is intended for general-purpose applications including semantic retrieval, classification, clustering, and cross-lingual text matching.

\emph{Multilingual MPNet}~\cite{reimers-gurevych-2020-making} is a multilingual sentence-transformer model trained on parallel and paraphrase data. It produces 768-dimensional sentence representations and is designed for semantic similarity, clustering, and semantic search across more than 50 languages.

\emph{LaBSE}~\cite{feng-etal-2022-language} was developed specifically for language-independent sentence representations and cross-lingual sentence matching. Its training objective encourages translated sentence pairs to occupy nearby regions of a shared multilingual embedding space.

\emph{Jina Embeddings v3}~\cite{sturua2024jina} is a multilingual multi-task
embedding model based on a Jina-XLM-RoBERTa architecture. It incorporates
Rotary Position Embeddings to support sequences of up to 8,192 tokens and
uses task-specific Low-Rank Adaptation (LoRA) adapters for applications
including retrieval, classification, clustering, and text matching. The model
also supports Matryoshka representations with a default embedding dimension
of 1,024.

The five models represent complementary multilingual embedding approaches:
BGE-M3 and Multilingual E5 are primarily retrieval-oriented, Multilingual
MPNet is paraphrase-oriented, LaBSE provides an established cross-lingual
sentence-alignment baseline, and Jina Embeddings v3 introduces a multi-task,
adapter-based embedding architecture. Their differing training objectives and
representation strategies make them suitable for comparing sensitivity to
controlled translation errors.

\section{Related Work}
\label{sec:related-work}

Previous work relevant to this study spans translation-error taxonomies,
contrastive challenge sets for metric evaluation, analyses of specific
translation phenomena, and multilingual evaluation resources.

Sharou and Specia~\cite{sharou-specia-2022-taxonomy} study
\textit{critical translation errors}, including incorrect dates, times,
numerical values, currencies, named entities, and other errors with potentially
serious consequences for users. Their work shows that translation errors differ
in severity and that evaluation should consider both their linguistic form and
their practical impact. Hayakawa and Arase~\cite{hayakawa-arase-2020-fine}
adopt a fine-grained error-analysis perspective for English-to-Japanese neural
machine translation in the medical domain, covering phenomena such as
addition, omission, mistranslation, grammatical errors, and terminology. These
studies motivate the use of linguistically meaningful error categories, but
their primary focus is the analysis or annotation of machine translation
output rather than the sensitivity of general-purpose sentence embeddings.

The methodology of the present work is more closely related to contrastive
challenge-set evaluation. ACES, introduced by Amrhein
et al.~\cite{amrhein-etal-2022-aces}, contains controlled correct--incorrect
translation pairs spanning 68 linguistic and semantic phenomena. A metric
succeeds when it assigns a higher score to the correct translation, allowing
category-specific weaknesses to be identified even when aggregate correlations
appear strong. Avramidis and
Macketanz~\cite{avramidis-macketanz-2022-linguistically} similarly use a
linguistically motivated challenge set for German--English and
English--German metric evaluation and show that strong metrics may still
struggle with named entities, terminology, measurement units, dates, idiomatic
expressions, and grammatical constructions. The present study adopts the same
contrastive principle but evaluates general-purpose multilingual sentence
encoders directly rather than complete machine translation metrics.

Dedicated neural MT evaluation metrics provide an important comparison for
general-purpose sentence embeddings. COMET~\cite{rei-etal-2020-comet}
uses pretrained cross-lingual representations to predict translation quality
from source, candidate, and reference information. Reference-free variants in
the COMET family, commonly referred to as COMETKiwi~\cite{rei-etal-2022-cometkiwi}, estimate translation quality directly from the source sentence and candidate translation without requiring a target-language reference. In the present study, COMETKiwi is used as an MT-specific quality-estimation baseline against which the contrastive sensitivity of the general-purpose embedding models can be compared.

More broadly, MT evaluation has been supported by a range of datasets and
metric families designed to capture different aspects of translation quality.
The WMT Metrics Shared Tasks provide system outputs paired with expert human
judgements, including MQM-based annotations, and support meta-evaluation of
automatic metrics at both system and segment level
~\cite{freitag-etal-2023-results}. Alongside contrastive resources such as
ACES, linguistically motivated challenge sets have been used to assess metric
sensitivity to phenomena including named entities, terminology, measurement
units, grammatical constructions, and discourse-level distinctions
~\cite{avramidis-etal-2023-challenging}. Contemporary evaluation approaches
include lexical metrics such as BLEU and chrF and learned neural metrics such
as COMET, COMETKiwi, and MetricX, with the latter families exploiting
pretrained multilingual representations and also supporting reference-free
quality-estimation variants
~\cite{rei-etal-2020-comet,rei-etal-2022-cometkiwi,
juraska-etal-2023-metricx}. This diversity motivates examining whether
general-purpose multilingual sentence-embedding similarity provides
complementary signals for fine-grained translation-error detection.

Specific linguistic phenomena further motivate the selected error categories.
Hossain et al.~\cite{hossain-etal-2020-non} show that negation is a persistent
source of machine translation error across language directions. Negation is
particularly relevant to embedding-based evaluation because a single lexical
change can reverse sentence truth conditions while leaving most of the surface
form unchanged. The same issue arises for numbers, dates, named entities,
measurement units, terminology, omissions, additions, modality, and semantic
roles: a translation may remain fluent and lexically similar to the reference
while differing in an important factual or semantic detail.

FLORES+ provides professionally translated and sentence-aligned multilingual data for machine translation evaluation. Gordeev et al.~\cite{gordeev-etal-2024-flores} illustrate its role as an extensible multilingual evaluation resource. In the present study, FLORES+ serves as the source of the English--Greek parallel sentences used to construct the contrastive dataset.

The resulting study differs from previous work in its object of evaluation.
Rather than assessing machine translation systems or task-specific evaluation
metrics, it examines whether multilingual sentence-embedding models preserve
fine-grained distinctions that are important for translation accuracy. By
combining controlled contrastive examples with category-level analysis, the
study evaluates not only whether a correct translation is ranked above an
erroneous alternative, but also how strongly the two are separated in the
embedding space. This provides an error-specific view of the strengths and
blind spots of multilingual sentence representations for English--Greek
translation evaluation.

\section{Methodology}
\label{sec:methodology}

For each English source sentence, the dataset contains a correct Greek
reference translation and a Greek translation containing one controlled error.
The evaluation is based on a cross-lingual, source-based comparison, in which
the English source and each Greek candidate translation are encoded with the
same multilingual sentence-embedding model.

For a model \(m\), consider and English sentence  \(s_{\mathrm{en}}\), a correct Greek translation \(t_{\mathrm{el}}^{+}\)
of  \(s_{\mathrm{en}}\) and a corresponding erroneous Greek translation  \(t_{\mathrm{el}}^{-}\).
For each contrastive example, two similarities scores are defined as follows:

\begin{equation}
S_{\mathrm{m}}^{+}
=
\cos\left(
E_m(s_{\mathrm{en}}),
E_m(t_{\mathrm{el}}^{+})
\right),
\label{eq:correct-source-score}
\end{equation}

and

\begin{equation}
S_{\mathrm{m}}^{-}
=
\cos\left(
E_m(s_{\mathrm{en}}),
E_m(t_{\mathrm{el}}^{-})
\right),
\label{eq:error-source-score}
\end{equation}

where \(E_m(\cdot)\) is the embedding function of model \(m\)

The \emph{contrastive margin} for model \(m\) is then defined as:

\begin{equation}
\Delta_m
=
S_{\mathrm{m}}^{+}
-
S_{\mathrm{m}}^{-}.
\label{eq:contrastive-margin}
\end{equation}

A positive value of \(\Delta_m\) indicates that the model assigns a higher
similarity score to the correct translation. A value close to zero indicates
limited sensitivity to the introduced error, whereas a negative value
indicates that the erroneous translation is preferred. The magnitude of the
margin reflects how clearly the model separates the two alternatives.

Model performance is evaluated using \emph{contrastive accuracy} and the \emph{mean
contrastive margin}. \emph{Contrastive accuracy} is defined as the proportion of
examples for which:

\begin{equation}
S_{\mathrm{m}}^{+}
>
S_{\mathrm{m}}^{-}.
\label{eq:contrastive-correct}
\end{equation}

To account for floating-point precision, score differences with an absolute contrastive margin of at most $10^{-7}$ were treated as numerical ties. Accordingly, an example was counted as correct when $\Delta_m > 10^{-7}$, incorrect when $\Delta_m < -10^{-7}$, and tied when $|\Delta_m| \leq 10^{-7}$. Ties were not counted as correct predictions when computing contrastive accuracy. The same numerical tolerance was applied to the COMETKiwi score margins.

For each error category \(c\), the \emph{mean contrastive margin} is calculated as:

\begin{equation}
\overline{\Delta}_{m,c}
=
\frac{1}{N_c}
\sum_{i=1}^{N_c}
\left(
S_{\mathrm{m},i}^{+}
-
S_{\mathrm{m},i}^{-}
\right),
\label{eq:mean-category-margin}
\end{equation}

where \(N_c\) denotes the number of examples in category \(c\). This
category-level measure is used to compare model sensitivity across different
types of translation errors.

The same evaluation procedure is applied independently to BGE-M3,
Multilingual E5, Multilingual MPNet, LaBSE, and Jina Embeddings v3,
allowing direct comparison of their overall ranking accuracy and their
category-specific sensitivity to controlled translation errors.

For the COMETKiwi baseline, the same contrastive ranking principle is used,
but the model-specific quality score replaces cosine similarity. For each
example, the English source sentence is paired separately with the correct and
erroneous Greek candidate translations. Let $Q^{+}$ and $Q^{-}$ denote the
corresponding COMETKiwi scores. The baseline succeeds on an example when

\begin{equation}
Q^{+} > Q^{-}.
\end{equation}

The corresponding COMETKiwi score margin is defined as

\begin{equation}
\Delta_{\mathrm{CK}} = Q^{+} - Q^{-}.
\end{equation}

This permits direct comparison of contrastive accuracy across methods. However,
COMETKiwi scores and cosine similarities are defined on different scales;
therefore, their raw margin magnitudes are not compared directly across
methods.

\section{Experimental Setup}
\label{sec:experimental-setup}

This section describes the implementation and configuration used for the two
classes of evaluated methods. We first report the checkpoints, encoding
settings, and preprocessing used for the five multilingual sentence-embedding
models. We then describe the configuration of the COMETKiwi
quality-estimation baseline used for the contrastive comparison.

\subsection{Sentence-Embedding Experimental Setup}
\label{subsec:embedding-setup}

All models were evaluated using the Sentence Transformers framework
(version 3.4.1), together with Transformers 4.48.3, Tokenizers 0.21.0,
and Hugging Face Hub 0.27.1. Table~\ref{tab:model-setup} reports the Hugging Face checkpoints, model parameter counts, and maximum input lengths used in the experiments conducted between August 5 and August 10, 2026.

\begin{table}[h!]
\caption{Embedding models and configurations used in the experiments.}
\label{tab:model-setup}
\centering
\scriptsize
\begin{tabular}{|l|l|r|r|}
\hline
\textbf{Model} &
\textbf{Checkpoint} &
\textbf{Parameters} &
\textbf{Max. Tokens} \\
\hline\hline

BGE-M3 &
\texttt{BAAI/bge-m3} &
560M &
8192 \\
\hline

Multilingual E5 &
\begin{tabular}[t]{@{}l@{}}
\texttt{intfloat/} \\
\texttt{multilingual-e5-large}
\end{tabular} &
560M &
512 \\
\hline

Multilingual MPNet &
\begin{tabular}[t]{@{}l@{}}
\texttt{sentence-transformers/} \\
\texttt{paraphrase-multilingual-mpnet-base-v2}
\end{tabular} &
278M &
128 \\
\hline

LaBSE &
\texttt{sentence-transformers/LaBSE} &
471M &
256 \\
\hline

Jina Embeddings v3 &
\texttt{jinaai/jina-embeddings-v3} &
570M &
8192 \\
\hline

\end{tabular}
\end{table}

Models were executed on a CUDA device with a batch size of 16. Sentence
representations were generated using the pooling configuration provided by each
Sentence Transformers checkpoint. Embeddings were L2-normalized during
encoding, and cosine similarity was computed as the dot product between
normalized vectors. The default truncation behaviour of each checkpoint was
retained.

For Multilingual E5, the \texttt{query:} prefix was added to the English
source sentence as well as to the correct and erroneous Greek translations.
BGE-M3 was used through the standard Sentence Transformers interface without
an additional instruction prompt, and its dense sentence representation was
used for similarity computation. For Jina Embeddings v3, the \texttt{text-matching} task adapter was used,
since the evaluation compares the semantic similarity of paired texts.
Embeddings were generated at the model's default dimensionality.

Greek sentences were normalized to Unicode NFC form and stripped of leading
and trailing whitespace before encoding. English sentences were stripped of
leading and trailing whitespace when loaded from the dataset. No additional
linguistic preprocessing was applied.

\subsection{COMETKiwi Baseline}
\label{subsec:cometkiwi}

As a dedicated machine translation evaluation baseline, we additionally
evaluated the reference-free COMETKiwi model~\cite{rei-etal-2022-cometkiwi}. For each contrastive example, the English sentence was provided as the source, and the correct and erroneous Greek translations were scored independently as candidate translations. Contrastive accuracy was computed using the same ranking criterion and numerical tie tolerance as for the embedding models, while the COMETKiwi margin was defined as the score of the
correct translation minus that of the erroneous translation. Because COMETKiwi
scores and cosine similarities operate on different numerical scales, their raw
margin magnitudes are not compared directly across methods.

For COMETKiwi experiments, \texttt{Unbabel/wmt22-cometkiwi-da} (repository revision updated in April 2025) was used.

\section{Dataset Construction}
\label{sec:dataset-construction}

This section describes the source data, the controlled error categories, and
the main statistics of the resulting English--Greek contrastive dataset.

\subsection{Source Data}
\label{subsec:source-data}

The contrastive dataset was constructed from the English--Greek portion of
FLORES+. FLORES+ provides sentence-aligned, human-produced reference
translations for multilingual machine translation evaluation. The benchmark
contains 2,009 distinct underlying sentences, comprising 997 sentences in the
development (\texttt{dev}) split and 1,012 sentences in the development-test
(\texttt{devtest}) split. Because the same underlying sentences are translated
across the supported languages, FLORES+ enables sentence-level multilingual
alignment and systematic cross-lingual evaluation.

In the present study, we use a subset of 1,212 aligned English--Greek
sentence pairs from this broader benchmark. These pairs form the source pool
from which the 1,850 contrastive examples were constructed. The number of
contrastive examples exceeds the number of unique sentence pairs because some
FLORES+ sentences are used to construct more than one controlled error variant
belonging to different error categories.

Each selected record consists of an English source sentence
\(s_{\mathrm{en}}\) and its corresponding Greek reference translation
\(t^{+}_{\mathrm{el}}\).

The FLORES+ translations are treated as high-quality reference translations
rather than as unique ideal translations, since a source sentence may admit
multiple valid translations. For each selected English--Greek pair, an
erroneous Greek variant \(t^{-}_{\mathrm{el}}\) was created by introducing
one intended translation error. The resulting contrastive unit is represented
as:

\begin{equation}
\left(
s_{\mathrm{en}},
t^{+}_{\mathrm{el}},
t^{-}_{\mathrm{el}},
c
\right),
\label{eq:contrastive-unit}
\end{equation}

where \(c\) denotes the assigned error category.

\subsection{Error Categories}
\label{subsec:error-categories}

The error inventory was informed by translation-accuracy phenomena studied in
ACES and by critical-error taxonomies. The dataset contains ten core error
categories and five exploratory categories. Core categories contain 150
examples each, whereas exploratory categories contain between 50 and 75
examples.

The contrastive dataset was constructed collaboratively by two translation experts. For each English source sentence, the experts examined the corresponding Greek FLORES+ reference translation and created an erroneous Greek variant containing one intended translation error. Errors involving explicit numbers, dates, negation, and lexical substitutions were introduced according to clear and reproducible transformation criteria. Semantic-role, coreference, terminology, and discourse-level errors received particular attention because modifications in these categories may affect grammaticality, alter unintended aspects of meaning, or introduce more than one error.

Each final erroneous translation was reviewed and approved by both experts to verify that it contained the intended error, belonged to the assigned category, differed semantically from the correct reference translation, and remained sufficiently fluent and natural in Greek. During dataset construction, the experts raised concerns about the optimality of the original Greek reference translation for 18 of the 1,212 underlying FLORES+ sentence pairs (1.5\%). In these cases, the original FLORES+ translation was retained unchanged in order to preserve consistency with the source dataset.

The selected categories cover factual, lexical-semantic, grammatical,
relational, referential, and discourse-level translation phenomena. Their
definitions and the number of examples included in each category are presented
in Table~\ref{tab:error-categories}

\begin{table}[h!]
\caption{Error categories used in the English--Greek contrastive dataset. $\mathcal{N}$ denotes the number of contrastive examples in each category.}
\label{tab:error-categories}
\centering
\scriptsize

\begin{tabular}{|
    p{0.20\linewidth}|
    p{0.10\linewidth}|
%    p{0.15\linewidth}|
    p{0.48\linewidth}|
    p{0.05\linewidth}|
}
\hline
\textbf{Category} &
\textbf{Type} &
\textbf{Meaning} &
\textbf{$\mathcal{N}$} \\
\hline\hline

Negation &
Core &
Changes the polarity or truth conditions of the sentence. &
150 \\
\hline
Number &
Core &
Changes a numerical value in the translation. &
150 \\
\hline
Named entity &
Core &
Replaces a person, place, organization, or other named entity. &
150 \\
\hline
Antonym &
Core &
Replaces a word with one expressing an opposite meaning. &
150 \\
\hline
Omission &
Core &
Removes information present in the source sentence. &
150 \\
\hline
Unsupported addition &
Core &
Introduces information that is absent from the source. &
150 \\
\hline
Hallucination &
Core &
Adds a complete unsupported proposition. &
150 \\
\hline
Tense and aspect &
Core &
Changes the temporal interpretation or completion of an event. &
150 \\
\hline
Modality &
Core &
Changes possibility, certainty, permission, or obligation. &
150 \\
\hline
Semantic role &
Core &
Reverses participant roles, such as agent and patient. &
150 \\
\hline
Date and time &
Exploratory &
Changes a date, year, time, or temporal expression. &
75 \\
\hline
Unit and currency &
Exploratory &
Changes a unit, currency, percentage, or measurement type. &
50 \\
\hline
Terminology &
Exploratory &
Replaces a domain-related term with an incorrect term. &
75 \\
\hline
Pronoun and coreference &
Exploratory &
Changes a pronoun, referent, gender, or number. &
75 \\
\hline
Discourse connective &
Exploratory &
Changes the logical relation between clauses. &
75 \\
\hline
\end{tabular}
\end{table}

\subsection{Dataset Statistics}
\label{subsec:dataset-statistics}

The final dataset contains 1,850 contrastive examples based on 1,212 unique
FLORES+ sentence pairs. A source sentence may occur in more than one error
category because different controlled error variants can be constructed from
the same reference translation.

The dataset contains 1,500 examples in the ten core categories and 350
examples in the five exploratory categories. The intended exploratory size was
75 examples per category; however, only 50 suitable unit- or currency-bearing
sentences were available in the selected FLORES+ data.

Table~\ref{tab:dataset-summary} summarizes the principal dataset statistics.

To support reproducibility, the complete contrastive dataset, including the
scores produced by the evaluated methods, as well as the code used to run the
experiments and compute the evaluation measures, is publicly available in a
GitHub repository\footnote{\url{https://github.com/DLIB-Ionian-University/translation-sentence-embeddings}}.

\begin{table}[h!]
\caption{Summary statistics of the generated contrastive
dataset.}
\label{tab:dataset-summary}
\centering
\scriptsize
\begin{tabular}{|l|r|}
\hline
\textbf{Statistic} & \textbf{Value} \\
\hline\hline
Total contrastive examples & 1,850 \\
\hline
Unique FLORES+ sentence pairs & 1,212 \\
\hline
Core error categories & 10 \\
\hline
Exploratory error categories & 5 \\
\hline
Core-category examples & 1,500 \\
\hline
Exploratory-category examples & 350 \\
\hline
%Examples from \texttt{dev} & 931 \\
%Examples from \texttt{devtest} & 919 \\
%Preliminarily labelled critical errors & 1,175 \\
%Preliminarily labelled major errors & 675 \\
%Human-validated examples at draft stage & 0 \\
Mean English source length & 22.64 words \\
\hline
Mean correct Greek translation length & 24.78 words \\
\hline
Mean erroneous Greek translation length & 25.23 words \\
\hline
\end{tabular}
\end{table}

\section{Experimental Results}
\label{sec:results}

This section reports the contrastive performance of the evaluated methods.
We first present the results of the five multilingual sentence-embedding
models, including overall and category-level contrastive accuracy and
within-model margins. We then compare their performance with the dedicated
reference-free COMETKiwi quality-estimation baseline using the same
contrastive ranking criterion.

\subsection{Sentence-Embedding Results}
\label{subsec:embedding-results}

Table~\ref{tab:overall-performance} summarizes the overall performance of
the five evaluated models. Table~\ref{tab:accuracy-by-category} reports
contrastive accuracy by error category, while
Table~\ref{tab:margin-by-category} reports the corresponding mean contrastive
margins defined in Eq.~\ref{eq:mean-category-margin}. Contrastive accuracy indicates how often the correct
translation is preferred, whereas the margin provides a model-specific measure
of the separation between the correct and erroneous alternatives.

\begin{table}[h!]
\caption{Overall contrastive performance of the evaluated models. Incorrect
denotes cases in which the erroneous translation received a higher similarity
score than the correct translation; ties denote cases in which both
translations received the same similarity score following the numerical tolerance defined in Section~\ref{sec:methodology}.\\}
\label{tab:overall-performance}
\centering
\scriptsize
\begin{tabular}{|l|r|r|r|r|r|}
\hline
\textbf{Model} &
\textbf{Acc. (\%)} &
\textbf{Mean $\Delta$} &
\textbf{Median $\Delta$} &
\textbf{Incorrect} &
\textbf{Ties} \\
\hline\hline
BGE-M3             & \textbf{89.30} & 0.0271 & 0.0155 & 197 & 1 \\
\hline
Multilingual E5    & 81.35 & 0.0095 & 0.0053 & 344 & 1 \\
\hline
Multilingual MPNet & 88.05 & 0.0387 & 0.0180 & 220 & 1 \\
\hline
LaBSE              & 87.73 & 0.0277 & 0.0198 & 226 & 1 \\
\hline
Jina Embeddings v3 & 87.73 & 0.0363 & 0.0213 & 226 & 1 \\
\hline
\end{tabular}
\end{table}

\begin{figure}[h!]
    \centering
    \includegraphics[width=0.82\textwidth]
%    {figures/overall_contrastive_accuracy.pdf}
    {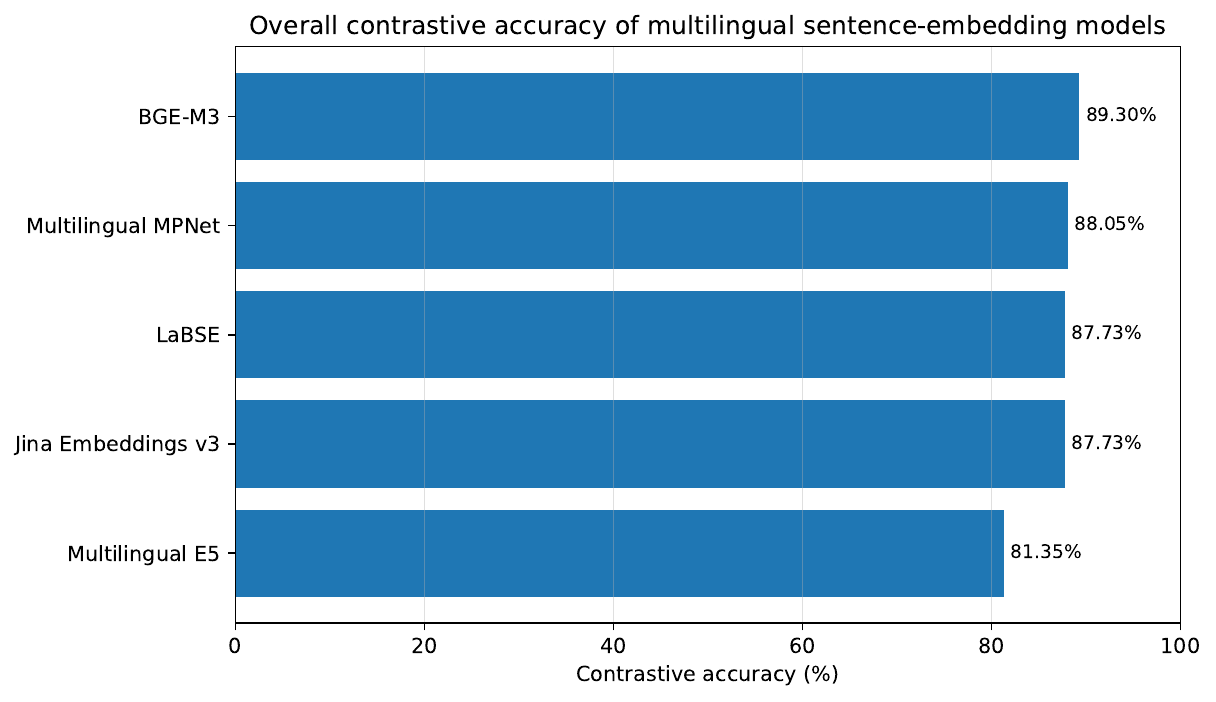}

    \caption{Overall contrastive accuracy of the five multilingual
    sentence-embedding models.}
    \label{fig:overall-accuracy}
\end{figure}

BGE-M3 achieved the highest contrastive accuracy at 89.30\%.
Multilingual MPNet obtained the largest raw mean contrastive margin
(0.0387), while Jina Embeddings v3 obtained the largest raw median margin
(0.0213). Jina Embeddings v3 and LaBSE achieved the same overall contrastive
accuracy of 87.73\%. Because cosine-similarity distributions may differ across
embedding models, cross-model differences in absolute margin magnitude should
be interpreted descriptively rather than as directly calibrated measures of
discrimination strength. Multilingual E5 showed the lowest overall accuracy
and the smallest raw mean and median margins. Figure~\ref{fig:overall-accuracy} provides a visual comparison of overall contrastive accuracy across the five sentence-embedding models.

\begin{table}[h!]
\caption{Contrastive accuracy (\%) by error category.}
\label{tab:accuracy-by-category}
\centering
\scriptsize
\begin{tabular}{|l|r|r|r|r|r|}
\hline
\textbf{Category} &
\textbf{BGE-M3} &
\textbf{Mult. E5} &
\textbf{Mult. MPNet} &
\textbf{LaBSE} &
\textbf{Jina Emb. v3} \\
\hline\hline
Negation                & 92.67 & 95.33 & 90.00 & 93.33 & 94.00\\
\hline
Number                  & 96.67 & 98.00 & 99.33 & 98.00 & 96.67\\
\hline
Named entity            & 99.33 & 93.33 & 99.33 & 95.33 & 98.00\\
\hline
Antonym                 & 86.67 & 80.00 & 89.33 & 92.00 & 85.33\\
\hline
Omission                & 84.00 & 81.33 & 92.67 & 92.67 & 86.00\\
\hline
Unsupported addition    & 95.33 & 91.33 & 90.67 & 97.33 & 91.33\\
\hline
Hallucination           & 99.33 & 100.00 & 99.33 & 85.33 & 100.00\\
\hline
Tense and aspect        & 64.00 & 47.33 & 69.33 & 71.33 & 61.33\\
\hline
Modality                & 93.33 & 79.33 & 87.33 & 96.00 & 90.00\\
\hline
Semantic role           & 88.00 & 70.00 & 77.33 & 69.33 & 87.33\\
\hline
Date and time           & 100.00 & 97.33 & 93.33 & 98.67 & 98.67\\
\hline
Unit and currency       & 86.00 & 86.00 & 84.00 & 84.00 & 76.00\\
\hline
Terminology             & 97.33 & 90.67 & 94.67 & 89.33 & 93.33\\
\hline
Pronoun and coreference & 61.33 & 41.33 & 60.00 & 40.00 & 60.00\\
\hline
Discourse connective    & 88.00 & 48.00 & 78.67 & 98.67 & 81.33\\
\hline
Overall                 & 89.30 & 81.35 & 88.05 & 87.73 & 87.73\\
\hline
\end{tabular}
\end{table}

Table~\ref{tab:accuracy-by-category} reveals substantial variation across error
categories and across models. Explicit factual and lexical changes were generally
detected reliably: numbers, named entities, hallucinations, and dates and times
produced accuracies above 95\% for several models. BGE-M3 performed particularly
well for named entities (99.33\%), hallucinations (99.33\%), and dates and times
(100\%), while also obtaining the highest accuracy for semantic-role errors
(88.00\%). Multilingual E5 achieved 100\% accuracy for hallucinations and strong
performance for numbers (98.00\%) and negation (95.33\%), but was substantially
weaker for tense and aspect (47.33\%), pronoun and coreference (41.33\%), and
discourse connectives (48.00\%). Multilingual MPNet achieved 99.33\% for both
numbers and named entities and performed strongly for omissions (92.67\%) and
terminology (94.67\%). LaBSE showed particular strength for unsupported additions
(97.33\%), modality (96.00\%), dates and times (98.67\%), and discourse
connectives (98.67\%). Jina Embeddings v3 achieved 100\% accuracy for
hallucinations, 98.00\% for named entities, and 98.67\% for dates and times,
although its accuracy decreased to 76.00\% for unit-and-currency errors.

Despite these model-specific strengths, tense and aspect errors and pronoun coreference errors remained difficult across the model set. Accuracy
for tense and aspect ranged from 47.33\% for Multilingual E5 to 71.33\% for
LaBSE, whereas pronoun and coreference ranged from 40.00\% for LaBSE to
61.33\% for BGE-M3. Discourse-connective errors showed particularly pronounced
model-specific variation, ranging from 48.00\% for Multilingual E5 to 98.67\%
for LaBSE. These results indicate that no single model is uniformly superior
across error categories and that sensitivity depends strongly on the linguistic
phenomenon being evaluated.

\begin{table}[h!]
\caption{Mean contrastive cosine-similarity margin by error category.}
\label{tab:margin-by-category}
\centering
\scriptsize
\begin{tabular}{|l|r|r|r|r|r|}
\hline
\textbf{Category} &
\textbf{BGE-M3} &
\textbf{Mult. E5} &
\textbf{Mult. MPNet} &
\textbf{LaBSE} &
\textbf{Jina Emb. v3} \\
\hline\hline
Negation                & 0.0412 & 0.0149 & 0.0666 & 0.0233 & 0.0483\\
\hline
Number                  & 0.0238 & 0.0128 & 0.0456 & 0.0481 & 0.0523\\
\hline
Named entity            & 0.0696 & 0.0190 & 0.1129 & 0.0560 & 0.0881\\
\hline
Antonym                 & 0.0217 & 0.0073 & 0.0315 & 0.0186 & 0.0285\\
\hline
Omission                & 0.0226 & 0.0077 & 0.0411 & 0.0375 & 0.0361\\
\hline
Unsupported addition    & 0.0175 & 0.0055 & 0.0206 & 0.0345 & 0.0205\\
\hline
Hallucination           & 0.0490 & 0.0288 & 0.0774 & 0.0378 & 0.0865\\
\hline
Tense and aspect        & 0.0032 & 0.0005 & 0.0053 & 0.0056 & 0.0051\\
\hline
Modality                & 0.0226 & 0.0063 & 0.0265 & 0.0226 & 0.0204\\
\hline
Semantic role           & 0.0283 & 0.0038 & 0.0153 & 0.0123 & 0.0193\\
\hline
Date and time           & 0.0228 & 0.0125 & 0.0315 & 0.0330 & 0.0388\\
\hline
Unit and currency       & 0.0230 & 0.0066 & 0.0156 & 0.0178 & 0.0193\\
\hline
Terminology             & 0.0234 & 0.0060 & 0.0200 & 0.0191 & 0.0223\\
\hline
Pronoun and coreference & 0.0013 & $-0.0001$ & 0.0015 & 0.0001 & 0.0020\\
\hline
Discourse connective    & 0.0062 & $-0.0009$ & 0.0058 & 0.0266 & 0.0087\\
\hline
Overall mean            & 0.0271 & 0.0095 & 0.0387 & 0.0277 & 0.0363\\
\hline
\end{tabular}
\end{table}

Table~\ref{tab:margin-by-category} shows clear differences in within-model
contrastive-margin patterns across error categories. Multilingual MPNet
produced the largest raw overall mean margin (0.0387), followed by Jina
Embeddings v3 (0.0363), LaBSE (0.0277), BGE-M3 (0.0271), and Multilingual
E5 (0.0095). Jina Embeddings v3 obtained the largest raw median margin
(0.0213), followed by LaBSE (0.0198), Multilingual MPNet (0.0180),
BGE-M3 (0.0155), and Multilingual E5 (0.0053).

Within each model, the largest margins were generally associated with explicit
factual or lexical changes. BGE-M3 produced relatively large margins for named
entities (0.0696), hallucinations (0.0490), and negation (0.0412).
Multilingual E5 exhibited much smaller margins overall, although its largest
values occurred for hallucinations (0.0288), named entities (0.0190), and
negation (0.0149). Multilingual MPNet showed particularly large margins for
named entities (0.1129), hallucinations (0.0774), negation (0.0666), and
numbers (0.0456). LaBSE produced comparatively large margins for named
entities (0.0560), numbers (0.0481), hallucinations (0.0378), omissions
(0.0375), unsupported additions (0.0345), and dates and times (0.0330).
Jina Embeddings v3 showed relatively large margins for named entities
(0.0881), hallucinations (0.0865), numbers (0.0523), and negation (0.0483).

Across the five models, the smallest margins were consistently observed for
tense-and-aspect and pronoun--coreference errors. For tense and aspect, mean
margins ranged from 0.0005 for Multilingual E5 to 0.0056 for LaBSE, while for
pronoun and coreference they ranged from $-0.0001$ for Multilingual E5 to
0.0020 for Jina Embeddings v3. Discourse-connective margins were also strongly
model-dependent, ranging from $-0.0009$ for Multilingual E5 to 0.0266 for
LaBSE.

Overall contrastive accuracy and raw contrastive margin provide complementary
descriptive information. BGE-M3 achieved the highest overall accuracy at
89.30\%, followed by Multilingual MPNet at 88.05\%, while LaBSE and Jina
Embeddings v3 both achieved 87.73\%. Multilingual E5 obtained the lowest
overall accuracy at 81.35\%. Because cosine-similarity distributions may differ
across embedding models, the absolute magnitude of raw margins should not be
interpreted as a directly calibrated cross-model measure of discrimination
strength. Raw margins are therefore most informative for examining relative
sensitivity patterns across error categories within each model.

\subsection{Comparison with COMETKiwi}
\label{subsec:cometkiwi-results}

As a dedicated MT quality-estimation baseline, COMETKiwi achieved an overall
contrastive accuracy of 94.49\%, exceeding the accuracies of all five
general-purpose sentence-embedding models. It correctly preferred the correct
translation in 1,748 of the 1,850 contrastive examples, with 102 incorrect
rankings and no ties. Its mean and median within-metric score margins were
0.0673 and 0.0441, respectively. These margin values are reported
descriptively for COMETKiwi and are not directly comparable with the raw
cosine-similarity margins of the embedding models.

Table~\ref{tab:cometkiwi-results} reports category-level COMETKiwi
performance. The metric achieved particularly high contrastive accuracy for
unsupported additions (100\%), negation (99.33\%), omissions (99.33\%),
semantic-role errors (99.33\%), hallucinations (98.67\%), named entities
(98.67\%), and modality (98.00\%). COMETKiwi also achieved
82.67\% accuracy for tense-and-aspect errors and 89.33\% for pronoun and
coreference, substantially above the corresponding ranges observed for the
five sentence-embedding models.

\begin{table}[h!]
\caption{Contrastive performance of COMETKiwi by error category. The margin
denotes the COMETKiwi score for the correct translation minus that for the
erroneous translation and is interpreted only within the COMETKiwi scoring
scale.}
\label{tab:cometkiwi-results}
\centering
\scriptsize
\begin{tabular}{|l|r|r|r|}
\hline
\textbf{Category} &
\textbf{$\mathcal{N}$} &
\textbf{Acc. (\%)} &
\textbf{Mean $\Delta_{\mathrm{CK}}$} \\
\hline\hline
Negation                & 150 & 99.33  & 0.1051 \\
\hline
Number                  & 150 & 89.33  & 0.0148 \\
\hline
Named entity            & 150 & 98.67  & 0.1130 \\
\hline
Antonym                 & 150 & 96.00  & 0.0400 \\
\hline
Omission                & 150 & 99.33  & 0.1198 \\
\hline
Unsupported addition    & 150 & 100.00 & 0.1103 \\
\hline
Hallucination           & 150 & 98.67  & 0.0639 \\
\hline
Tense and aspect        & 150 & 82.67  & 0.0112 \\
\hline
Modality                & 150 & 98.00  & 0.0366 \\
\hline
Semantic role           & 150 & 99.33  & 0.1563 \\
\hline
Date and time           & 75  & 72.00  & 0.0027 \\
\hline
Unit and currency       & 50  & 90.00  & 0.0618 \\
\hline
Terminology             & 75  & 93.33  & 0.0367 \\
\hline
Pronoun and coreference & 75  & 89.33  & 0.0132 \\
\hline
Discourse connective    & 75  & 93.33  & 0.0233 \\
\hline
Overall                  & 1850 & 94.49 & 0.0673 \\
\hline
\end{tabular}
\end{table}

COMETKiwi was not uniformly superior across categories. Its accuracy for
date-and-time errors was 72.00\%, compared with 93.33--100\% across the
sentence-embedding models. Its accuracy for number errors was 89.33\%, also
below all five embedding models. These differences indicate that dedicated MT
quality estimation and general-purpose multilingual sentence embeddings
exhibit complementary error-sensitivity profiles.

\section{Discussion}
\label{sec:discussion}

The results indicate that multilingual sentence embeddings provide useful
signals for translation evaluation, but their sensitivity depends strongly on
the type of translation error. Across the five evaluated models, explicit
factual and lexical changes were generally distinguished more reliably than
localized grammatical and referential errors. In particular, named entities,
hallucinations, numbers, and dates tended to produce relatively large
within-model contrastive margins and high contrastive accuracies. This pattern
was observed across several architectures: BGE-M3, Multilingual MPNet, and
Jina Embeddings v3 showed strong sensitivity to named-entity and hallucination
errors, while numbers and dates were detected reliably by most models. These
findings suggest that changes affecting salient semantic content are more likely
to alter the position of a sentence in the embedding space.

In contrast, tense-and-aspect and pronoun--coreference errors remained
difficult to trace across all five models. For tense and aspect, contrastive accuracy
ranged from 47.33\% for Multilingual E5 to 71.33\% for LaBSE, with Jina
Embeddings v3 obtaining 61.33\%. Pronoun and coreference produced an even
narrower separation between correct and erroneous translations, with
accuracies ranging from 40.00\% for LaBSE to 61.33\% for BGE-M3; Jina
Embeddings v3 achieved 60.00\%. The corresponding within-model margins were
also consistently small, including 0.0051 for tense and aspect and 0.0020 for
pronoun and coreference with Jina Embeddings v3. These phenomena may change
the validity of a translation without substantially altering its lexical content
or overall topic. Sentence-level embeddings therefore appear less sensitive to
certain fine-grained morphosyntactic and referential distinctions than to broad
semantic content. The consistency of this pattern across models with different
training objectives suggests that reference tracking and temporal distinctions
represent persistent blind spots for embedding-based translation evaluation.

The behaviour of discourse-connective errors was more model-dependent.
Multilingual E5 showed the weakest contrastive accuracy at 48.00\%, whereas
LaBSE achieved 98.67\%. BGE-M3, Multilingual MPNet, and Jina Embeddings v3
obtained intermediate accuracies of 88.00\%, 78.67\%, and 81.33\%,
respectively. A similar pattern was observed in the within-model margins:
Multilingual E5 produced a slightly negative mean margin ($-0.0009$), BGE-M3,
Multilingual MPNet, and Jina Embeddings v3 produced relatively small positive
margins of 0.0062, 0.0058, and 0.0087, respectively, whereas LaBSE produced
a noticeably larger raw margin of 0.0266. One possible explanation is that
LaBSE's cross-lingual sentence-alignment objective preserves some clause-level
semantic relations more effectively. This result should, however, be
interpreted cautiously, since performance within a single error category does
not establish general discourse understanding.

Contrastive accuracy and contrastive margin capture different aspects of model
behaviour. BGE-M3 achieved the highest overall contrastive accuracy at
89.30\%, followed by Multilingual MPNet at 88.05\%, while LaBSE and Jina
Embeddings v3 both achieved 87.73\%. Multilingual E5 obtained the lowest
overall accuracy at 81.35\%. The ranking based on raw margins was different:
Multilingual MPNet produced the largest raw mean contrastive margin (0.0387),
followed by Jina Embeddings v3 (0.0363), while Jina Embeddings v3 produced
the largest raw median margin (0.0213). This difference illustrates that the
frequency with which a model ranks the correct translation first and the
magnitude of the separation between alternatives need not yield the same model
ordering. However, because different embedding models may exhibit different
cosine-similarity distributions, the absolute magnitude of raw margins should
not be interpreted as a directly calibrated cross-model measure of
discrimination strength. Raw margins are therefore most informative for
examining relative sensitivity patterns within a given embedding space and
across error categories for the same model.

The results of Multilingual E5 further illustrate the limitations of relying on
absolute cosine similarity. Correct and erroneous translations were frequently
mapped to highly similar regions of its embedding space, resulting in the
smallest raw overall mean margin (0.0095) and median margin (0.0053) among
the evaluated models. Its particularly small or negative margins for
pronoun--coreference and discourse-connective errors show that high
cross-lingual similarity does not necessarily imply sensitivity to fine-grained
translation errors. Conversely, the comparatively larger margins produced by
Multilingual MPNet and Jina Embeddings v3 did not translate into the highest
overall contrastive accuracy, further demonstrating that raw similarity
separation and ranking consistency capture complementary properties.

Jina Embeddings v3 further shows that the observed category-level limitations
are not restricted to the other evaluated embedding architectures. Despite
achieving strong performance for hallucinations (100\%), named entities
(98.00\%), dates and times (98.67\%), and numbers (96.67\%), its performance
decreased substantially for tense-and-aspect errors (61.33\%), pronoun and
coreference (60.00\%), and unit-and-currency errors (76.00\%). Its overall
contrastive accuracy of 87.73\% was competitive with the other embedding
models, but its sensitivity profile remained strongly category-dependent. This
result reinforces the broader finding that differences in general-purpose
multilingual embedding architectures do not necessarily eliminate weaknesses
in the representation of localized grammatical and referential distinctions.

The comparison with COMETKiwi provides an important qualification to these
embedding-specific findings. COMETKiwi achieved substantially higher overall
contrastive accuracy (94.49\%) than any of the five general-purpose embedding
models. The difference was particularly pronounced for several phenomena that
were difficult for the embeddings. For tense-and-aspect errors, COMETKiwi
achieved 82.67\% accuracy, compared with 47.33--71.33\% across the embedding
models, while for pronoun and coreference it achieved 89.33\%, compared with
40.00--61.33\%. COMETKiwi also achieved 99.33\% accuracy for semantic-role
errors. These results suggest that an MT-specific quality-estimation objective
can preserve or exploit distinctions that are only weakly reflected in
general-purpose sentence-level embedding similarity.

At the same time, the comparison does not indicate uniform superiority of
COMETKiwi. Date-and-time errors provide a particularly clear counterexample:
COMETKiwi achieved only 72.00\% contrastive accuracy, whereas all five
sentence-embedding models exceeded 93\%. COMETKiwi was also less accurate
for number errors than any of the embedding models. This complementary pattern
suggests that general-purpose embedding similarity and MT-specific quality
estimation capture partly different aspects of translation adequacy. It also
supports category-level evaluation, since a substantially higher aggregate
accuracy can coexist with pronounced weaknesses for particular error types.

One limitation concerns dependence among contrastive examples. The 1,850
examples are derived from 1,212 unique FLORES+ sentence pairs, and some source
sentences occur in more than one error category. Consequently, the observations
are not fully independent. This does not affect the descriptive comparisons
reported here, but it should be taken into account in future inferential
analyses. In particular, confidence intervals or significance tests should use
resampling procedures clustered by the underlying FLORES+ sentence pair rather
than treating each corrupted example as an independent observation.

A second limitation concerns the textual domain represented by FLORES+.
The dataset primarily consists of general-domain, topic-diverse informational prose derived from Wikimedia sources, as opposed to domain-specific, conversational, or literary text. Consequently, extending similar experiments to literary and poetic texts would be valuable, as these genres exhibit different linguistic and stylistic properties, including ambiguity, lexical creativity, a greater use of figurative language and consistent explorations of form and rhythm. Such an evaluation could help determine whether the findings observed here may be generalized beyond the scope informational prose, and be applicable to more stylistically complex translation settings.

A further limitation concerns the scope of the baseline comparison. Although
the study includes COMETKiwi as a dedicated reference-free MT
quality estimation baseline, it does not yet include reference-based neural
metrics or simple lexical baselines. COMETKiwi should therefore not be treated
as representative of all MT evaluation approaches. Future work should broaden
the comparison to additional MT-specific and lexical metrics and examine
category-level agreement and complementary error sensitivity across methods.

Taken together, the findings suggest that general-purpose multilingual
sentence embeddings are better suited as components of broader
translation evaluation systems than as standalone metrics. They provide useful
semantic signals for explicit factual and lexical changes, while COMETKiwi's
higher overall contrastive accuracy indicates the value of MT-specific
quality-estimation signals for several grammatical, referential, and relational
phenomena. At the same time, COMETKiwi was less reliable for some explicit
factual categories, particularly date-and-time and number errors. This
complementary behaviour suggests that combining semantic, lexical, syntactic,
referential, and MT-specific signals may provide more robust coverage of
translation errors than relying on a single scoring approach.

Finally, the results show that no single method is uniformly superior across
all error categories. Among the sentence-embedding models, BGE-M3 provided
the highest overall ranking accuracy, Multilingual MPNet produced the largest
raw mean margin, Jina Embeddings v3 produced the largest raw median margin,
and LaBSE exhibited particular strength for several categories, including
discourse connectives. COMETKiwi achieved the highest overall contrastive
accuracy, but also showed pronounced weaknesses for specific error types.
The different sensitivity profiles observed across the five embedding models
and COMETKiwi therefore support category-level evaluation rather than reliance
on a single aggregate score.

\section{Conclusion and Future Work}
\label{sec:conclusion}

This study evaluated the ability of five multilingual sentence-embedding
models---BGE-M3, Multilingual E5, Multilingual MPNet, LaBSE, and Jina
Embeddings v3---to detect controlled translation errors in English--Greek
contrastive pairs, and compared their behaviour with a dedicated reference-free
COMETKiwi quality-estimation baseline. Among the embedding models, BGE-M3
achieved the highest contrastive accuracy at 89.30\%, Multilingual MPNet
produced the largest raw mean margin, and Jina Embeddings v3 produced the
largest raw median margin while matching LaBSE in overall accuracy. Across
the five embeddings, tense-and-aspect and pronoun--coreference errors remained
among the most difficult categories.

COMETKiwi achieved a higher overall contrastive accuracy of 94.49\% and
substantially improved detection of tense-and-aspect, pronoun--coreference,
and semantic-role errors. However, it was considerably less reliable for
date-and-time errors and also underperformed the embedding models for numbers.
The resulting category-specific differences show that general-purpose sentence
embeddings and MT-specific quality estimation provide complementary signals
rather than a uniformly ordered set of alternatives. Overall, the findings
support the use of multilingual sentence embeddings as components of broader
translation-evaluation frameworks rather than as standalone metrics.

Future work will extend the contrastive dataset to additional language pairs,
initially French--Greek and Italian--Greek, using the same error categories
and evaluation procedure. This will allow direct comparison of model
sensitivity across languages and help determine whether the observed strengths
and blind spots are language-specific or consistent across multilingual
settings. In addition, future experiments will broaden the MT-evaluation baseline set
beyond COMETKiwi by incorporating reference-based neural metrics and simple
lexical baselines. Particular attention will be given to correlations,
category-level agreement, and complementary error sensitivity across
general-purpose embeddings, MT-specific quality-estimation models, and
reference-based metrics. This will help determine whether combinations of
these signals can provide more robust coverage of factual, grammatical,
referential, relational, and discourse-level translation errors.

As a further extension, the scores and contrastive margins produced by the evaluated embedding models and MT-specific evaluation methods will be combined as input features for supervised machine-learning experiments. The aim will be to investigate whether classifiers can exploit the complementary information captured by these signals to improve the detection of translation errors, both overall and across individual error categories. This will allow us to examine whether combinations of model-based similarity and quality-estimation features provide more robust error detection than any single metric considered independently.


%\bibliographystyle{plain}
%\bibliography{bibliography}

\end{document}